\documentclass[journal,twoside,web]{ieeecolor}

\pdfoutput=1

\usepackage{colortbl}
\usepackage[table]{xcolor}
\usepackage{tmi}
\usepackage{cite}
\usepackage{amsmath,amssymb,amsfonts}
\usepackage{algorithmic}
\usepackage{textcomp}
\usepackage{adjustbox}
\usepackage{booktabs}
\usepackage{multirow}
\usepackage{arydshln}
\usepackage{pifont}
\usepackage{makecell}
\usepackage{xcolor}
\usepackage{graphicx}
\usepackage{subcaption}
\usepackage{pgfplots}

\makeatletter
\let\NAT@parse\undefined
\makeatother
\definecolor{tmiblue}{RGB}{0,126,215}
\usepackage[colorlinks=true,linkcolor=tmiblue,citecolor=tmiblue,urlcolor=tmiblue]{hyperref}
\usepackage[normalem]{ulem}

\def\etal{\emph{et al.}}

\def\ie{\emph{i.e.\ }}

\newcommand{\response}[1]{{\color[RGB]{0,0,0}{#1}}}

\def\BibTeX{{\rm B\kern-.05em{\sc i\kern-.025em b}\kern-.08em
    T\kern-.1667em\lower.7ex\hbox{E}\kern-.125emX}}
\begin{document}

\title{MultiEYE: Dataset and Benchmark for OCT-Enhanced Retinal Disease Recognition from Fundus Images}
\author{Lehan Wang, Chubin Ou, Chongchong Qi, Lin An, Mei Jin, Xiangbin Kong, Xiaomeng Li
\thanks{Lehan Wang is with the Department of Electronic and Computer Engineering at the Hong Kong University of Science and Technology, Hong Kong SAR, China (e-mail: \href{mailto:lwangdk@connect.ust.hk}{lwangdk@connect.ust.hk}).
Chubin Ou and Lin An are with the Guangdong Weiren Meditech Co., Ltd., Foshan, China (e-mail: \href{mailto:cou@connect.ust.hk}{cou@connect.ust.hk}; \href{mailto:lynnanncn@hotmail.com}{lynnanncn@hotmail.com}).
Chongchong Qi is with the Yunnan United Vision Innovations Technology Co., Ltd., Kunming, P. R. China (e-mail: \href{mailto:ccqi@uvi.ltd}{ccqi@uvi.ltd}).
Mei Jin is with the Department of Ophthalmology at the Guangdong Hospital of Integrated Traditional Chinese and Western Medicine, Foshan, China (e-mail: \href{mailto:jinmei75@163.com}{jinmei75@163.com}).
Xiangbin Kong is with the Department of Ophthalmology at the Second People's Hospital of Foshan, Foshan, China (e-mail: \href{mailto:xiangbin_kong@sina.com}{xiangbin\_kong@sina.com}).
Xiaomeng Li is with the Department of Electronic and Computer Engineering at the Hong Kong University of Science and Technology, Hong Kong SAR, China (e-mail: \href{mailto:eexmli@ust.hk}{eexmli@ust.hk}). Corresponding author: Xiaomeng Li.}
\thanks{This work is supported by grants from Foshan HKUST Projects (Grants FSUST21-HKUST10E and FSUST21-HKUST11E) and a grant from the Guangdong Provincial Science and Technology Fund (Project 2023A0505030004).}
}

\maketitle
\begin{abstract}
Existing multi-modal learning methods on fundus and OCT images mostly require both modalities to be available and strictly paired for training and testing, which appears less practical in clinical scenarios. To expand the scope of clinical applications, we formulate a novel setting, ``OCT-enhanced disease recognition from fundus images", that allows for the use of unpaired multi-modal data during the training phase, and relies on the widespread fundus photographs for testing.
To benchmark this setting, we present the first large multi-modal multi-class dataset for eye disease diagnosis, MultiEYE, 
and propose an OCT-assisted Conceptual Distillation Approach (OCT-CoDA), which employs semantically rich concepts to extract disease-related knowledge from OCT images and leverages them into the fundus model.
Specifically, we regard the image-concept relation as a link to distill useful knowledge from OCT teacher model to fundus student model, which considerably improves the diagnostic performance based on fundus images and formulates the cross-modal knowledge transfer into an explainable process. Through extensive experiments on the multi-disease classification task, our proposed OCT-CoDA demonstrates remarkable results and interpretability, showing great potential for clinical application. Our dataset and code will be made available at \href{https://github.com/xmed-lab/MultiEYE}{https://github.com/xmed-lab/MultiEYE}.
\end{abstract}

\begin{IEEEkeywords}
Knowledge Distillation, OCT-Enhanced Retinal Disease Classification, Vision Language Model
\end{IEEEkeywords}

\section{Introduction}
\label{sec:introduction}

\IEEEPARstart{F}{undus} photography and Optical Coherence Tomography (OCT) scanning are widely recognized as essential tools in the screening of eye diseases. Fundus photos offer a comprehensive view of the eye, capturing vital structures such as the retina, optic nerve, and vascular network~\cite{bernardes2011digital}. On the other hand, OCT images are highly sensitive in detecting abnormalities in retinal structure and thickness~\cite{hassan2015review}.
In clinical practice, doctors often rely on both fundus photos and OCT scans for accurate disease diagnosis~\cite{watanabe2022combining,hao2022value}. For instance, to detect epiretinal membrane, ophthalmologists first perform a fundus examination to assess the potential presence of a fibrous membrane in front of the macula.
If the membrane is suspected, they confirm the diagnosis and assess macular thickness with the OCT image. This process exemplifies the critical role of utilizing both modalities and the reliability of combining them.
In light of this, our goal is to exploit the unique strengths of each modality and harness their combined potential for disease recognition.


\begin{figure*}[t]
	\centering
	\includegraphics[width=0.85\textwidth]{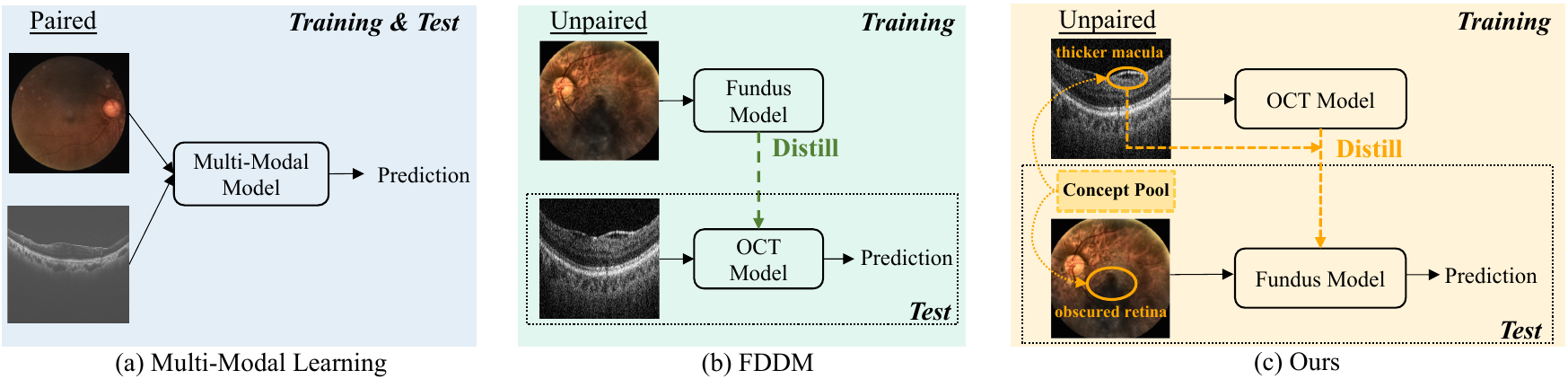}
	\caption{The comparison between multi-modal learning, FDDM~\cite{wang2023fundus} and our method. (a) Multi-modal learning requires paired data for both training and test stage. (b) FDDM and (c) our method similarly apply unpaired data for training and only utilize one modality for testing. In our method, we target at the fundus model and integrate concepts as additional guidance to help transfer teacher-dominant features to student model.}
	\label{fig:setting}
\end{figure*}

While several studies have explored multi-modal learning approaches~\cite{yoo2019possibility,wang2019two,li2021multi,li2022multimodal,wang2022learning} for OCT and fundus images, there are still substantial limitations.
\response{Firstly, these methods necessitate paired multi-modal data during both training and the inference stage, as shown in Figure~\ref{fig:setting}. However, public multi-modal datasets only contain a small scale of paired data and limited disease types due to the challenges of collecting such data, leaving a large number of public sources unused. FUND-OCT~\cite{hassan2022composite} only includes 4 types of diseases from 105 subjects, and MMC-AMD~\cite{wang2022learning} focuses merely on the categorization of Age-related Macular Degeneration (AMD) with fewer than 2000 images. This imposes high requirements for model training, limiting their broader adoption due to the scarcity of paired data. Moreover, existing methods adopted multimodal fusion to share knowledge between patient-level paired modalities, but the potential of using disease-level paired OCT and fundus images--where the images are not paired by patient--for training models has yet to be explored.}

\response{The work in~\cite{wang2023fundus} first investigated knowledge transfer between unpaired multimodal data by proposing a fundus-enhanced disease-aware distillation method for retinal disease classification from OCT images; see Figure~\ref{fig:setting} (b). 
However, it is more valuable to prioritize the fundus model and employ the OCT counterpart for additional guidance, due to \textbf{\textit{the higher prevalence of fundus images}} and \textbf{\textit{the additional depth information in OCT scans to enhance the fundus model}}.
Fundus cameras have a 5 times higher adoption rate than OCT devices, owing to the earlier invention and lower price~\cite{chopra2021optical}. Plenty of hospitals and clinics are equipped with only fundus cameras, making the amount of available fundus data larger than OCT scans. Even in hospitals that have both devices,  most patients typically undergo fundus photography first during examinations because it is cost-effective and provides a comprehensive overview of the eyes. OCT test is conducted when a condition cannot be diagnosed with fundus imaging, or the lesion needs further detailed evaluation. Thus, fundus photographs are more widely accessible compared with OCT scans.
Moreover, OCT scans can provide a cross-sectional view of the posterior vitreous, the vitreoretinal interface, the retina, and the choroid, capturing depth information about microscopic changes that are not visible in fundus photographs~\cite{muller2019ophthalmic,midena2020optical,davis2008comparison}. 
For these reasons, we follow~\cite{wang2023fundus} in training with unpaired data, but prioritize the fundus model instead. This setting, which we refer to as ``OCT-enhanced disease recognition from fundus images'', has several strengths:  (1) it is not constrained by the scarcity of paired multimodal datasets, allowing to leverage the abundant unpaired resources; (2) it capitalizes on the widespread fundus photographs, enabling the fundus model to inference with inherent knowledge learned from the OCT modality.}

In this paper, we collect a comprehensive unpaired multimodal multi-disease dataset called \textbf{MultiEYE}. This dataset comprises fundus and OCT images that share the same disease label space. Notably, the fundus and OCT images do not necessarily come from the same patient.
\response{Our key motivation is that although fundus photos do not provide direct observation of retinal layer structures, the model could learn the relation between distinct appearances across two modalities during training, and imply the corresponding cross-sectional structures from fundus features during inference. For example, it is challenging to determine whether the yellowish spots in a fundus photograph are drusen (indicative of AMD) or exudates (suggestive of other retinal conditions like DR) when relying solely on fundus photos. However, drusen and exudates are located in different layers, which can be differentiated by OCT scans. If the model can associate fundus image features of different diseases with different structure information in OCT, it will be more reliable in making such decisions, leading to a more accurate disease assessment.}

Therefore, it is crucial to determine \textit{\textbf{how to extract disease-related knowledge from OCT images}} and \textit{\textbf{how to effectively transfer these advantageous features to another modality}}. To achieve this, we integrate descriptive concepts to decouple helpful features in the OCT modality and serve as a linkage to convey knowledge between both modalities.
On this basis, we propose an \textbf{OCT}-enhanced \textbf{Co}nceptual \textbf{D}istillation \textbf{A}pproach (\textbf{OCT-CoDA}), as depicted in Figure~\ref{fig:setting} (c). Rather than utilizing a naive distillation, we perform distillation with the guidance of proposed concepts. Specifically, our method prompts a powerful Large Language Model to generate candidate concepts describing the specific symptoms for each disease, and adopts a Vision Language Model to obtain the similarity between concept embeddings and image features. Then, in order to transfer the concept similarity from teacher to student model at a disease level, we propose two types of knowledge distillation loss, global prototypical distillation, which aligns the general disease-concept relationships between both modalities, and local contrastive distillation, which narrows the concept similarities of samples from the same class.
Our contributions can be concluded as:
\begin{itemize}
    \item We establish a new setting, ``OCT-enhanced disease recognition from fundus images", for a more practical cross-modal retinal disease diagnosis scenario, and construct a multi-modal multi-class benchmark, MultiEYE, with 58,036 fundus photographs and 45,923 OCT B-scans. The dataset will be released for public access.
    \item We propose a conceptual distillation approach to enhance the performance of the fundus model with the OCT knowledge. Introducing fine-grained concepts as the connection between modalities ensures the knowledge distillation process more controllable and interpretable.
    \item We validate our method through extensive experiments on the MultiEYE dataset, and prove its efficiency as well as explainability over both single-modal and multi-modal networks.
\end{itemize}

\section{Related Works}
\label{sec:related-works}
\subsection{Multi-Modal Retinal Disease Diagnosis}

\begin{figure*}[t]
	\centering
	\includegraphics[width=0.85\textwidth]{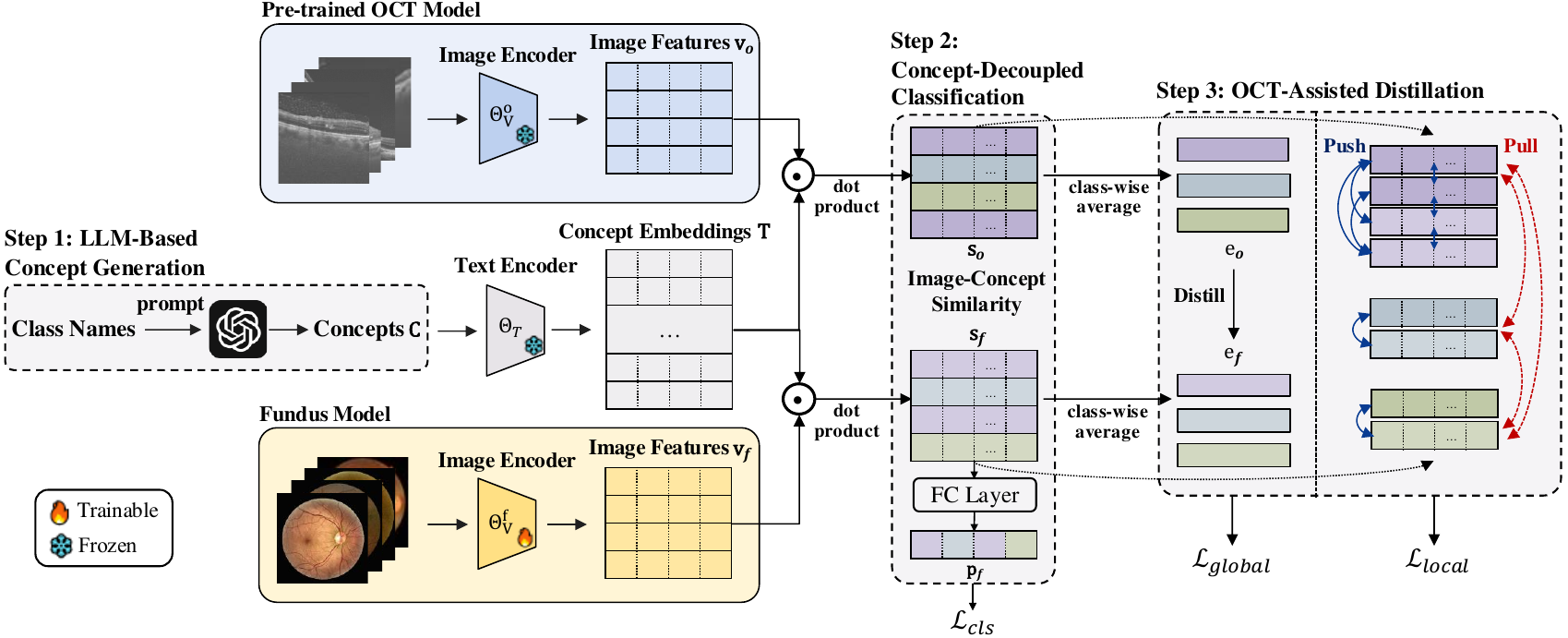}
	\caption{\response{The Framework of the Proposed OCT-CoDA Method. The pre-trained OCT model is adopted as the teacher to train the fundus student model. Given a batch of unpaired OCT images and fundus photos, they are fed into separate image encoders to get the extracted features. To implement the conceptual distillation, we first prompt the LLM to generate a concept pool. Secondly, we compute the similarity between image features and concept embeddings for each modality. Finally, the OCT-assisted distillation is performed based on the image-concept similarity. In the inference stage, we input this similarity matrix into a Fully Connected (FC) layer to obtain the prediction score.}}
	\label{fig:overview}
\end{figure*}

Previous studies have proved that Artificial Intelligence models are capable of achieving remarkable results in diagnosis by fusing multi-modal fundus and OCT images. 
Yoo \etal~\cite{yoo2019possibility} first attempted to explore AMD detection with paired fundus and OCT images, which demonstrated the effectiveness of combined modalities. Built on this, Wang \etal~\cite{wang2019two,wang2022learning} extended the multi-modal AMD classification setting to a two-stream structure which separately extracted fundus and OCT features and adopted the concatenation for prediction. He \etal~\cite{he2021multi} designed modality-specific attention networks based on the distinct features of two modalities respectively. 
Later on, more attention has been focused on the improvement of fusion strategies. Li \etal~\cite{li2021multi} introduced a multi-instance learning approach that fuses fundus instances with sequences of OCT B-scans through the attention mechanism. Li \etal~\cite{li2022multimodal} created a hierarchical fusion algorithm to effectively fuse 2D fundus images with 3D OCT volumes for glaucoma and diabetic retinopathy classification. However, these methods depend on the availability of both modalities during both training and test process, and the multi-modal data must be precisely matched at the patient level. 

Thus, the work in~\cite{wang2023fundus} enhanced the OCT modality by utilizing unpaired images during training and only required OCT scans for testing. Nevertheless, this setting has limited feasibility in widespread application due to the relatively high costs of OCT imaging. Besides, this method integrates the multi-modal knowledge without abundant guidance, which may introduce noise from teacher modality and lack interpretability on what kind of knowledge is transmitted between modalities.
To tackle this limitation, our purpose is to promote the classification performance with widely accessible fundus photos for broader clinical adoption. Moreover, we introduce disease concepts as the medium for distillation, making the knowledge transfer process more explainable and controllable.

\subsection{Large Language Models \& Vision Language Models}
In recent years, Large Language Models (LLMs) have witnessed significant advances and symbolized a revolutionary change in an extensive range of natural language processing tasks, exemplified by 
Generative Pre-trained Transformer (GPT) series~\cite{brown2020language,openai2023gpt4}. 
Meanwhile, the rise of Vision-Language Models (VLMs) bridge the gap between image and text. Pre-trained with large-scale image-text pairs, VLMs successfully capture abundant image-text correspondences.
For example, Contrastive Language-Image Pre-Training (CLIP)~\cite{radford2021learning} formed a shared embedding space for image and text through the contrastive learning method. In the field of ophthalmology, Silva \etal~\cite{silva2023foundation} pre-trained a foundation VLM for fundus image understanding by incorporating expert domain knowledge expressed in textual prompts. 

Recently, research has explored harnessing the knowledge obtained from LLMs to strengthen VLMs. Works in~\cite{menon2022visual,pratt2023does,feng2023text,liu2023chatgpt} leveraged the rich contextual knowledge inherent in LLMs to generate descriptive sentences for image categories, which promoted the effectiveness and generalizability of VLMs. 
Researches in ~\cite{yang2023language,oikarinen2023label,han2023llms} acquired a large number of concepts from LLM and designed interpretable bottleneck models for image classification based on CLIP, relieving the burden of manual concept annotations. 
The essence of LLM-aided concept bottleneck models was also extended to the medical domain in ~\cite{kim2023concept,byra2023few,yan2023robust} to build robust and interpretable medical image classifiers. 
In this paper, we harness powerful LLM to extract disease symptoms. Hence, disease features can be decoupled into fine-grained concepts, serving as a crucial link between two distinct image modalities.

\subsection{Cross-Modal Distillation}

Knowledge distillation\cite{hinton2015distilling} has been applied to multi-modal scenarios with the aim of transferring knowledge from superior modalities to inferior modalities. In the general domain, researches commonly performed knowledge distillation between visual RGB images and depth images~\cite{gupta2016cross}, texts~\cite{passalis2018learning}, or acoustics~\cite{chen2021distilling}.
In the medical domain, cross-modal distillation is particularly prevailing in Computed Tomography (CT) or Magnetic Resonance Imaging (MRI) scan segmentation. For instance, Jiang \etal~\cite{jiang2021unpaired} instructed a student CT network with an informative teacher MRI network. 
Chen \etal~\cite{chen2021learning} adopted a pre-trained multi-modal MRI model as the teacher to transfer privileged knowledge to a uni-modal MRI model. 
Built upon previous researches, Song \etal~\cite{song2022asynchronous} applied the cross-modal distillation to ophthalmology by transferring knowledge from a multi-modal OCT and VF network to a single-modal OCT network to enhance its capability in detecting glaucoma. 

Inspired by the idea of cross-modal distillation, we similarly apply the knowledge distillation algorithm to assist fundus modality with OCT scans. Differently, we separate disease attributes into fine-grained concepts and execute disease-specific knowledge distillation through these concepts.

\section{Methodology}
\label{sec:methodology}
An overview of our proposed OCT-CoDA framework is presented in Figure~\ref{fig:overview}. Our primary objective is to facilitate retinal disease diagnosis on fundus modality with the assistance of OCT modality. Initially, we establish a teacher model to extract features from the OCT modality. Subsequently, we implement knowledge distillation to strengthen the OCT dominant features in the fundus student model. Directly distilling features would bring less informative knowledge, such as background noise or disease-irrelevant features. Therefore, we design a novel conceptual distillation approach to introduce additional guidance. 
In this section, we first introduce the LLM-based concept generation process (\ref{sec:gpt}), and then illustrate the basic structure of the concept-decoupled network for classification (\ref{sec:concept}). Next, we describe our proposed OCT-assisted distillation approach (\ref{sec:distill}), which is composed of global prototypical distillation and local contrastive distillation. 

\subsection{LLM-Based Concept Generation}
\label{sec:gpt}

\begin{figure}[t]
	\centering
	\includegraphics[width=0.49\textwidth]{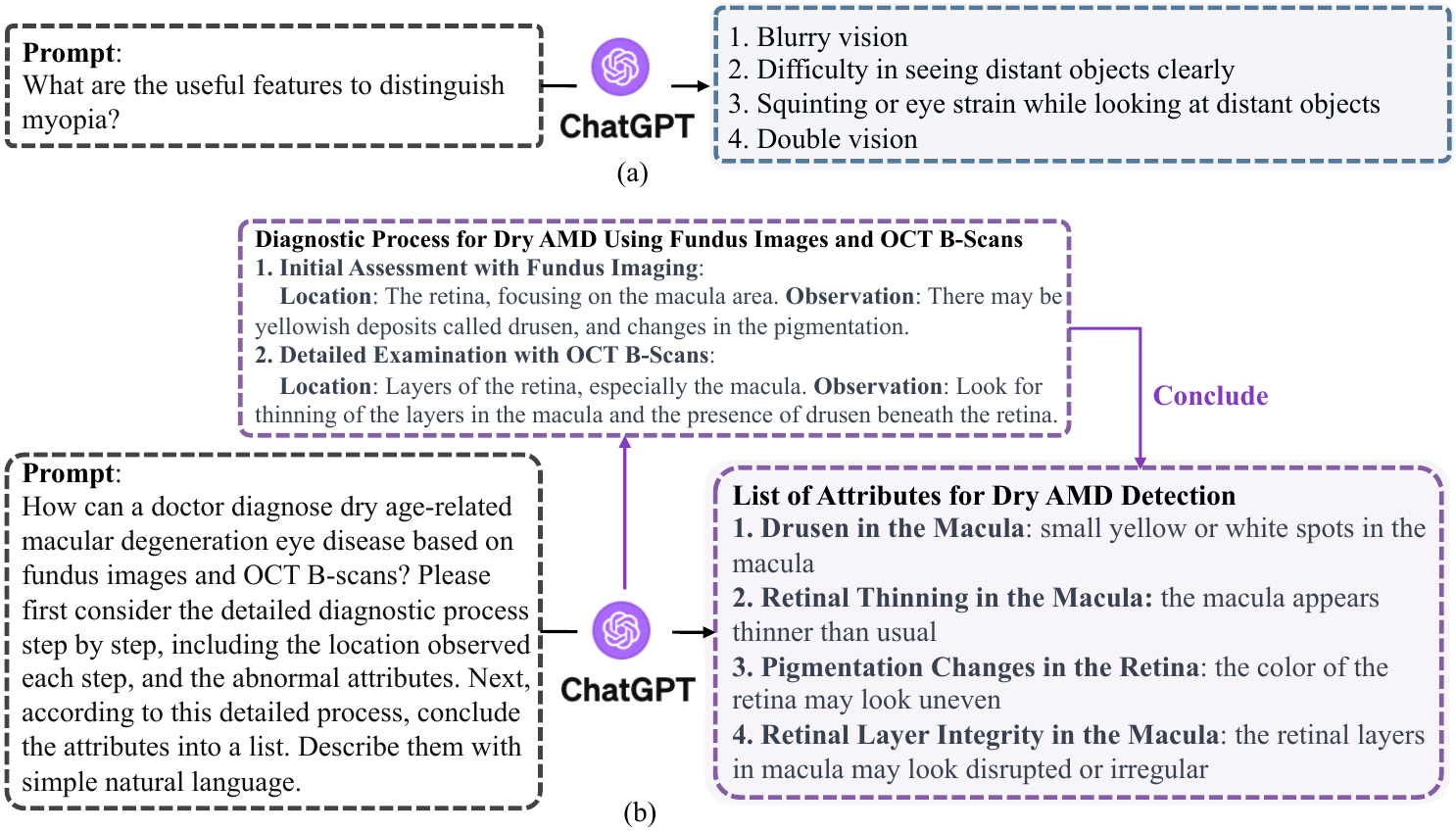}
	\caption{Examples of the concept generation process with LLM. (a) The typical prompt is used to trigger the LLM. (b) We apply our refined CoT prompt for concept generation.}
	\label{fig:cot}
\end{figure}

The first step of our method is to acquire a candidate set of concepts that thoroughly characterize the specific symptoms of each eye disease. However, manually annotating attributes for each category is exhausting. 
Given that powerful LLMs have sufficient knowledge in healthcare, we adopt GPT-4~\cite{openai2023gpt4} to autonomously generate attributes for each disease. A typical prompt choice could be \textit{``What are the useful features to distinguish \{disease name\}"}. However, such prompts are suboptimal in medical scenarios, as LLM might omit the precise location of the abnormalities or generate general symptoms invisible in the images such as ``blurry vision", as shown in Figure~\ref{fig:cot} (a).

To address this challenge, we follow the idea of Chain-of-Thought (CoT) prompting~\cite{wei2022chain}. 
In light of the VLM's sensitivity to location information, we instruct GPT-4 to diagnose diseases by examining different regions step by step, mimicking the behavior of professional doctors. An example of the generation procedure is shown in Figure~\ref{fig:cot} (b). Initially, GPT-4 examines fundus photographs to formulate a preliminary diagnosis, followed by a detailed analysis using OCT B-scans. Ultimately, we ask GPT-4 to conclude the involved information into a comprehensive list, which contains abnormal changes in color, shape as well as lesion regions.

\subsection{Concept-Decoupled Classification}
\label{sec:concept}
VLMs have shown a strong ability to align images with texts in a common space. As a consequence, we leverage VLM to ground the LLM-generated concepts by calculating the similarity between the image features and the text embeddings. Through this, the model can better relate images from different modalities with visible symptoms and illustrate their contributions respectively.

More specifically, let $\Theta_V$ and $\Theta_T$ denote the visual and text encoder in the VLM. 
Each encoder comprises a feature extractor and a projection head.
After obtaining a set of $N$ concepts $\bold{C}=\{c_i\}_{i=1}^N$, we extract text embeddings for concept $c_i$ and image features for image $x_j$ as follows:
\begin{align}
     t_i &= \Theta_T(c_i), i=1,...,N\:, \\
     v_j &= \Theta_V(x_j), j=1,...,B\:,
\end{align}
where $t_i$ is embedding of the $i^{th}$ concept in the concept set, and $v_j$ is the feature of the $j^{th}$ image in the batch of $B$ samples.

Given image representation $v_j$ and the set of concept embeddings $\bold{T}=\{t_i\}_{i=1}^N$, we can determine the similarity between the image and the concepts by computing their cosine distance, which is denoted as:
\begin{equation}
    \bold{s}=cos(v_j,\bold{T})=v_j \cdot \bold{T}^\top\:,
\end{equation}

\response{The relations between eye diseases and symptoms is more complex. To illustrate, different eye diseases may exhibit same symptoms while instances with the same disease may also show different sets of symptoms. For example, both retinal vein occlusion (RVO) and Diabetic Retinopathy (DR) might be related to various combinations of “hard exudates”, “retinal hemorrhages” and “cotton wool spots”. However, since the severity and distribution of these signs may vary depending on different diseases, we employ a concept classifier, denoted as $\Phi$, to learn the complicated disease-concept relationships from the image-concept similarity and output the final prediction.}

The prediction process can be described as:
\begin{equation}
    p=Softmax(\Phi(\bold{s}))\:,
\end{equation}
where $p$ is the final prediction. We denote the vision encoder with the concept classifier as $\Theta=[\Theta_V, \Phi]$.
     
\subsection{OCT-Assisted Distillation}
\label{sec:distill}
Based on the architecture above, our goal is to train a fundus classification network assisted by a pre-trained OCT model with the multi-modal data. In the inference stage, we discard the OCT branch and only leave the fundus model for prediction.
We freeze the text encoder $\Theta_T$, and build two visual encoders with concept classifiers, $\Theta^f$ and $\Theta^o$, for fundus and OCT modality respectively. $\Theta^f$ is our target fundus classification model. \response{At first, we pre-train the teacher model $\Theta^o=[\Theta_V^o, \Phi^o]$ on OCT images. Following the process described in Section~\ref{sec:concept}, we can obtain the prediction of OCT model $p_o$. Cross-entropy loss is applied as the classification loss for optimization, denoted as $\mathcal{L}_{pretrain} = -\sum_{i=1}^N{y_{o,i}\log{p_{o,i}}}$.}
To utilize knowledge from the OCT modality during training, we further freeze the OCT teacher model and train the fundus student model with the multi-modal training set. 
To explicitly transfer the modality-specific knowledge from the OCT teacher model to the fundus student model, we design two types of knowledge distillation losses, namely, {global prototypical distillation, which aligns disease-concept relationships across both modalities, as well as local contrastive distillation, which pulls closer the concept similarities for samples within the same class.}

\subsubsection{Global Prototypical Distillation (GPD)}
It is noteworthy that each disease is corresponded with a series of typical attributes, which can be reflected by the general similarity between the predefined concepts and the features of samples from this disease. 
Our purpose is to align the global concept-disease similarity of fundus and OCT modalities.
To illustrate, we compress the concept similarity into a prototype for each class, which represents the comprehensive characteristics of the disease.
During the training per batch, the class prototype is calculated as the average of all the concept similarity scores of images in this category, which is formulated as:
\begin{equation}
    \bold{e}^d_f=\frac{\sum_{i=1}^B{s_{f,i}*y_{f,i}^d}}{\sum_{i=1}^B{y_{f,i}^d}}, \ \bold{e}^d_o=\frac{\sum_{j=1}^B{s_{o,i}*y_{o,j}^d}}{\sum_{j=1}^B{y_{o,j}^d}} \:,
\end{equation}
where $\bold{e}^d_f$ and $\bold{e}^d_o$ denote the class $d$ prototype of the fundus and OCT modality respectively, $s_{f,i}$ $\left( s_{o,i}\right)$ represents the similarity score between the input image and the predefined concepts, and $y_{f,i}^d$ $\left(y_{o,j}^d\right)$ is a binary number which indicates whether the instance belongs to class $d$ or not.
Then, we define the global prototypical distillation loss as the distance between class prototypes:
\begin{equation}
    \mathcal{L}_{global}=\| \bold{e}^d_o - \bold{e}^d_f \|_2^2 \:,
\end{equation}
Through the global prototypical distillation, the generalized knowledge of relationships between disease and concepts could be transferred from the OCT model to the fundus counterpart. Thus, the fundus model would be particularly improved in the ability to identify less evident disease features in fundus photographs.

\subsubsection{Local Contrastive Distillation (LCD)}
While global prototypical distillation facilitates the OCT teacher model to impart generalized disease-level knowledge to the fundus student model, this approach might overlook sample-level details that equally contribute to disease recognition. Simultaneously, it is essential for the fundus modality to maintain its distinctive features, rather than relying entirely on the OCT modality.
To address this problem, we further propose local contrastive distillation to guide the fundus modality at the sample level. 
For an individual sample, the instances from the same class are considered as positives, while those from different categories are treated as negatives. Significantly, fundus photographs and OCT images are grouped together as positives if they belong to the same disease. Through this, the fundus model can not only learn the sample-concept relationships from the OCT model, but also preserve the distinct features in the original fundus modality by drawing closer fundus image representations of the same class. The process can be formulated as:
\begin{equation}
    \mathcal{L}_{local} = \sum_{i=1}^B{-\frac{1}{P_i}\sum_{p \in P_i}{\log\frac{\exp({s_{f,i}\cdot s_p/\tau)}}{\sum_{q \neq p}{\exp({s_{f,i}\cdot s_q/\tau)}}}}} \:,
\end{equation}
where $P_i$ denotes the set of all the positives for the $i^{th}$ sample, which contains both fundus images and OCT scans from the same class with $i$, and $\tau$ is the temperature scale that controls the smoothing strength. By implementing local contrastive distillation, the fundus model is encouraged to learn sample-wise disease features from the OCT modality, and concurrently preserve its unique modality-specific advantages. This dual focus ensures a balanced enhancement of the fundus model's diagnostic capabilities, integrating external knowledge without hindering its inherent strengths.

\subsubsection{Total Loss Function}
The total loss function for training the fundus student model is the summation of the classification loss $\mathcal{L}_{cls}$ and the distillation loss $\mathcal{L}_{distill}$. The classification loss is defined as cross-entropy loss, and the distillation loss is composed of global prototypical distillation and local contrastive distillation.
\begin{equation}
    \begin{aligned}
    \mathcal{L}&= \mathcal{L}_{cls} + \mathcal{L}_{distill}\:, \\
    &= \mathcal{L}_{cls} + \alpha\mathcal{L}_{global} + \beta\mathcal{L}_{local}\:,
\end{aligned}
\end{equation}
where $\alpha$ and $\beta$ are the weights that control the contribution of each distillation
loss.


\begin{table*}[t]
    \caption{\response{Sub-dataset and device distribution in the constructed MultiETE dataset.}}
    \label{tab:sub-dataset}
    \centering
    \begin{adjustbox}{width=0.8\textwidth}
        \begin{tabular}{llccc}
        \toprule
         \textbf{Dataset} & \textbf{Device} & \textbf{Train} & \textbf{Val} & \textbf{Test} \\ 
         \midrule
         RFMiD & TOPCON 3D OCT-2000, Kowa VX-10$\alpha$, TOPCON TRC-NW300 & 848 & 278 & 319 \\
         RFMiD 2.0 & TOPCON TRC-NW300, CARL ZEISS FF450 & 228 & 83 & 84 \\
         MMC-AMD & TOPCON fundus camera & 560 & 154 & 81 \\
         Messidor-2 & Topcon TRC NW6 non-mydriatic fundus camera & - & 181 & 1563 \\
         JSIEC&ZEISS FF450 Plus IR Fundus Camera, Topcon TRC-50DX Mydriatic Retinal Camera & 75 & - & 242 \\ 
         ODIR-5K & Various cameras including Canon, Zeiss and Kowa & 5585 & 844 & 1604 \\
         STARE & TopCon TRV-50 fundus camera & 286 & - & - \\
         VietAI & Unknown & 1251 & - & 282 \\
         ARIA & Zeiss FF450+ fundus camera & 134 & - & - \\ 
         FIVES & TRC-NW8 fundus cameras & 800 & - & -\\
         FUND-OCT & Topcon 3D OCT 2000 & 111 & 48 & - \\ 
         OIA-DDR & Topcon TRC NW6 & - & 5800 & 1034 \\ 
         EYEPACS & Various cameras including Canon CR2AF, NFC and Pictor Prestige & 24535 & 4207 & 6384 \\ 
         In-house & Topcon DRI Triton Swept Source OCT and Velite C3000 Spetral Domain OCT & 421 & 6 & 8 \\
         \bottomrule
        \end{tabular}
    \end{adjustbox}
    \vspace{0.05cm} \\
    (a) Sub-dataset distribution of fundus data in MultiEYE \\
    \vspace{0.15cm}
    \begin{adjustbox}{width=0.7\textwidth}
        \begin{tabular}{llcc}
        \toprule
         \textbf{Dataset} & \textbf{Device} & \textbf{Train} & \textbf{Test} \\ 
         \midrule
         OCT2017 & Spectralis OCT, Heidelberg Engineering, Germany & 29538 & 8018 \\
         OCTID & Cirrus HD-OCT machine & 470 & - \\
         MMC-AMD & Topcon OCT camera and Heidelberg OCT camera & 730 & 179 \\
         GOALS & TOPCON DRI Swept Source OCT & 85 & 15 \\
         In-house & Topcon DRI Triton Swept Source OCT and Velite C3000 Spetral Domain OCT & 5900 & 988 \\
         \bottomrule
        \end{tabular}
    \end{adjustbox}
    \vspace{0.05cm} \\
    (b) Sub-dataset distribution of OCT data in MultiEYE \\
\end{table*}

\section{Experiments}
\label{sec:experiments}

\subsection{Datasets}
\begin{figure*}[t]
	\centering
	\includegraphics[width=0.85\textwidth]{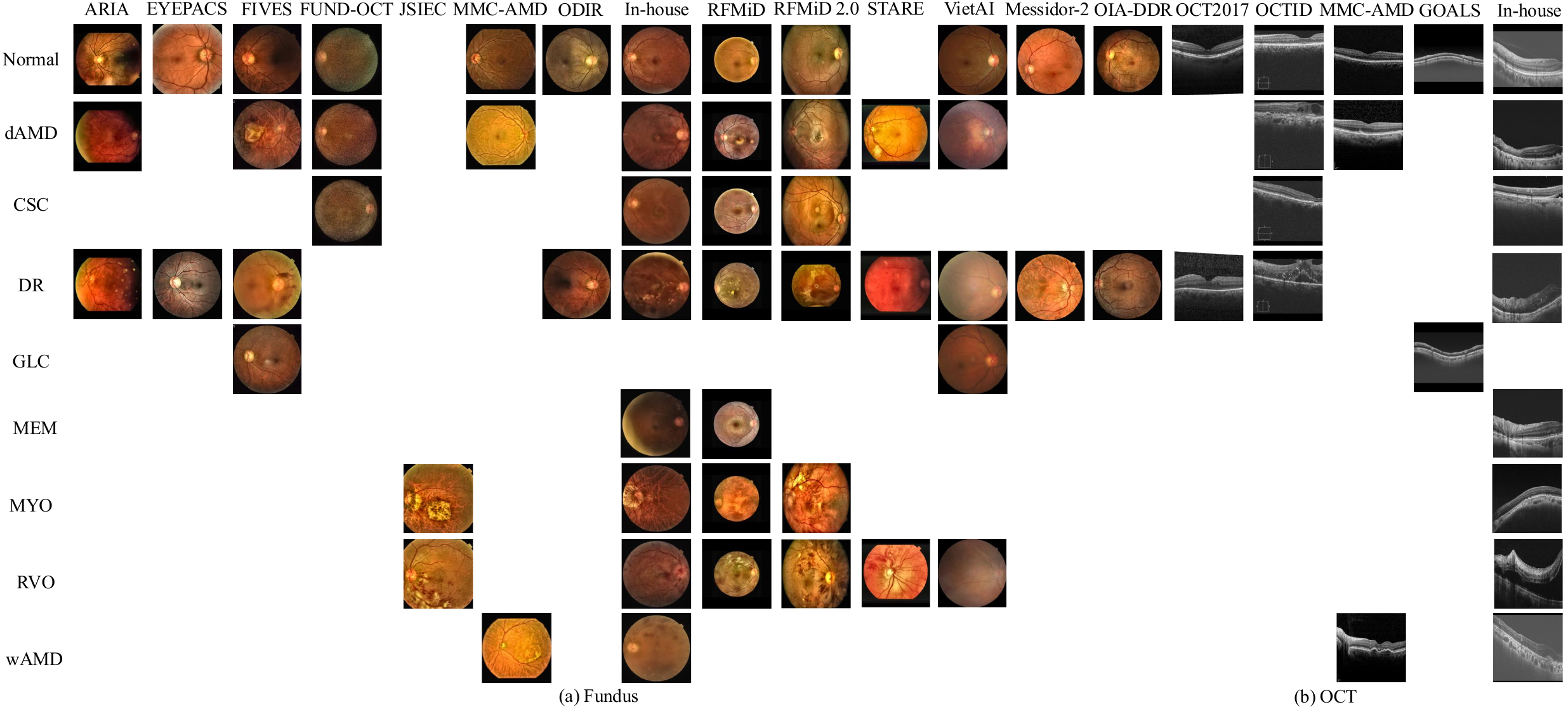}
	\caption{\response{Examples of the MultiEYE dataset. We select one sample for each category in each sub-dataset.}}
	\label{fig:data-sample}
\end{figure*}

To create a multi-modal multi-disease classification dataset, MultiEYE, we assemble 13 public fundus datasets and 4 OCT datasets with our in-house data. \response{The sub-dataset and device distributions are shown in Table~\ref{tab:sub-dataset}}.
For the public datasets, we include RFMiD1.0~\cite{s3g7-st65-20}, RFMiD2.0~\cite{mrd2-ap11-23}, MMC-AMD~\cite{wang2022learning}, Messidor-2\footnote{https://www.adcis.net/en/third-party/messidor2/}, 1000Fundus39Cat~\cite{cen2021automatic}, ODIR-5K\footnote{https://odir2019.grand-challenge.org/}, STARE\footnote{https://cecas.clemson.edu/ahoover/stare/}, VietAI\footnote{https://www.kaggle.com/c/vietai-advance-retinal-disease-detection-2020/data}, ARIA~\cite{farnell2008enhancement}, FIVES~\cite{jin2022fives}, FUND-OCT~\cite{hassan2022composite}, OIA-DDR\cite{li2019diagnostic}) and part of EyePACS\footnote{https://www.kaggle.com/c/diabetic-retinopathy-detection} in the fundus datasets; and OCT2017\footnote{https://www.kaggle.com/datasets/paultimothymooney/kermany2018}, OCTID~\cite{gholami2020octid}, MMC-AMD~\cite{wang2022learning} and GOALS~\cite{fang2022dataset} in the OCT datasets.
\response{All public datasets comprise central OCT B-scans that were selected based on clear disease indicators from the full volumetric scans.}
Our in-house data originates from the Guangdong Hospital of Integrated Traditional Chinese and Western Medicine, collected with Topcon DRI Triton Swept Source OCT and Velite C3000 Spetral Domain OCT. 
\response{Following the construction process of these public OCT datasets, we have similarly preprocessed our in-house OCT data by selecting B-scans that show signs of disease from the entire volume.}
All the cases were examined by two ophthalmologists independently to determine the diagnostic label.
\response{Among all the annotations made by junior ophthalmologists, there are approximately 30\% in disagreement with each other or marked as uncertain. These would be sent to senior ophthalmologists with over 15 years of experience to determine the final diagnosis.}
\response{We split the dataset based on patient identity, ensuring that no images from the same patient appear in the same subset. We maintained a training-validation-test split ratio of approximately 6:2:2.} As shown in the statistics, we extract nine classes, including normal, dry age-related macular degeneration (dAMD), central serous chorioretinopathy (CSC), diabetic retinopathy (DR), glaucoma (GLC), macular epiretinal membrane (MEM), myopia (MYO), retinal vein occlusion (RVO), and wet age-related macular degeneration (wAMD). \response{Examples of each class from each sub-dataset can be found in Figure~\ref{fig:data-sample}}. A summary of our MultiEYE dataset is presented in Table~\ref{tab:data-statistics}. 

\begin{table*}[t]
    \caption{Data statistics of the MultiEYE dataset.}
    \label{tab:data-statistics}
    \centering
    \begin{adjustbox}{width=0.75\textwidth}
        \begin{tabular}{ccccccccccc}
        \hline
         \textbf{Category} & Normal & dAMD & CSC & DR & GLC & MEM & MYO & RVO & wAMD & Total \\ 
         \hline
         \multicolumn{11}{c}{\textit{Fundus Dataset}} \\
         \hline
         \textbf{Train} &  22905 & 1167 & 94 & 8787 & 626 & 160 & 327 & 367 & 401 & 34834 \\
         \textbf{Val} & 7217 & 84 & 12 & 3939 & 134 & 13 & 57 & 26 & 119 & 11601 \\
         \textbf{Test} & 6960 & 139 & 30 & 3932 & 138 & 42 & 101 & 98 & 161 & 11601 \\
        \hline
        \multicolumn{11}{c}{\textit{OCT Dataset}} \\
        \hline
        \textbf{Train} &  22146 & 173 & 1014 & 10447 & 35 & 1281 & 183 & 795 & 649 & 36723 \\
        \textbf{Test} &  5708 & 64 & 186 & 2570 & 15 & 304 & 20 & 258 & 75 & 9200 \\
        \hline
        \end{tabular}
    \end{adjustbox}
\end{table*}

\subsection{Implementation Details}
In our experiments, we test our method with two backbones, CLIP~\cite{radford2021learning} and FLAIR~\cite{silva2023foundation}. We use ResNet-50 as the encoder and follow the original setting of each backbone to resize all images to the same size, namely, 224$\times$224 for CLIP and 512$\times$512 for FLAIR. Zero-padding is applied to rectangular images to avoid distortions. Following prior works~\cite{wang2019two,wang2022learning,wang2023fundus}, we use contrast-limited adaptive histogram equalization for fundus images and median filter for OCT images in pre-processing. Also, we adopt data augmentation including random crop, flip, rotation, and changes in contrast, saturation, and brightness. For a fair comparison, we apply identical data processing steps, data augmentation operations, model backbones, and running epochs in all the experiments. We use AdamW to optimize parameters with a learning rate of $1e^{-4}$, and cosine annealing for learning rate decay. The batch size is set to 64. For weight hyperparameters, $\tau$ is set to 10, $\alpha$ is set to 0.6 and $\beta$ is set to 0.05. All the models are implemented on an NVIDIA GeForce RTX 3090 GPU.

\subsection{Evaluation Metrics}
\response{
We utilize 8 metrics, \ie Precision, Recall, Specificity, Precision-Recall F1 score, Sensitivity-Specificity F1 score, Mean Average Precision, Accuracy and Kappa to evaluate the diagnosis performance.
For the $i^{th}$ class, we define the number of true positive, false positive, true negative, and false negative test samples as $TP_i$, $FP_i$, $TN_i$, $FN_i$. The computation of performance metrics is as follows:
\begin{gather*}
    Precision_i=\tfrac{TP_i}{TP_i+FP_i},Recall_i=\tfrac{TP_i}{TP_i+FN_i},\\
    P-R \ F1_i=\tfrac{2*Precision_i*Recall_i}{Precision_i+Recall_i} \\
\end{gather*}
\begin{gather*}
    Specificity_i=\tfrac{TN_i}{TN_i+FP_i}, Sensitivity_i = \tfrac{TP_i}{TP_i+FN_i}, \\
S-S \ F1_i=\tfrac{2*Sensitivity_i*Specificity_i}{Sensitivity_i+Specificity_i}, \\
Accuracy_i=\tfrac{TP_i+TN_i}{TP_i+FP_i+FN_i+TN_i},  \\
Kappa_i=\tfrac{2*(TP_i*TN_i-FP_i*FN_i)}{(TP_i+FP_i )*(FP_i+TN_i)+(TP_i+FN_i)*(FN_i+TN_i)},
\end{gather*}
Average Precision (AP) is the area under the precision-recall curve, and mean Average Precision (mAP) is calculated by finding Average Precision (AP) for each class and then average over all the classes. Our metrics can also provide a comprehensive performance evaluation even in the class-imbalanced setting, since true positive and true negative samples are generally incorporated in two kinds of F1 scores and mAP considers the precision-recall tradeoff across different thresholds.}

\begin{table*}[t]
    \caption{Performance comparison on the MultiEYE dataset. We select the best model on the validation set and report the average result on the test set.}
    \label{tab:main-result}
    \centering
    \begin{adjustbox}{width=0.74\textwidth}
        \begin{tabular}{l|cccccccc}
        \hline
         \textbf{Method} & \textbf{Precision} & \textbf{Recall} & \textbf{Specificity} & \textbf{P-R F1} & \textbf{S-S F1} & \textbf{MAP} &  \textbf{Accuracy} & \textbf{Kappa} \\ 
         \hline
         \multicolumn{9}{c}{\textit{CLIP Backbone}} \\
         \hdashline
         {Fundus} & 63.69 & 56.91 & 95.00 & 56.35 & 65.76 & 59.80 & 78.01 & 56.63 \\
         {Fundus+Concept} & 62.68 & 57.64 & 94.96 & 56.90 & 67.09 & 59.21 & 78.32 & 56.81 \\
         FDDM~\cite{wang2023fundus} & \textbf{67.81} & 55.67 & 94.97 & 58.21 & 65.67 & 59.01  & 77.98 & 56.49 \\
         \rowcolor{cyan!60!blue!20}
         Ours & 66.17 & \textbf{58.30} & \textbf{95.10} & \textbf{59.12} & \textbf{67.39} & \textbf{60.99} & \textbf{78.54} & \textbf{57.71} \\
         \hline
         \multicolumn{9}{c}{\textit{FLAIR Backbone}} \\
         \hdashline
         {Fundus} & 62.29 & 59.24 & 95.28 & 57.74 & 68.79 & 62.50 & 78.29 & 58.08 \\ 
         {Fundus+Concept} & 60.73 & 60.79 & 95.32 & 57.76 & 70.30 & 61.52  & 78.89 & 59.25 \\
         FDDM~\cite{wang2023fundus} & 65.44 & 59.27 & 95.29 & 58.79 & 67.04 & 61.90 & 79.23 & 59.19 \\
         \rowcolor{cyan!60!blue!20}
         Ours & \textbf{67.15} & \textbf{62.61} & \textbf{95.37} & \textbf{62.02} & \textbf{72.25} & \textbf{64.37} & \textbf{79.34} & \textbf{59.67} \\ 
         \hline
        \end{tabular}
    \end{adjustbox}
    \vspace{0.1cm} \\
    (a) We evaluate baselines and our method with 2 different backbones, CLIP~\cite{radford2021learning} and FLAIR~\cite{silva2023foundation}. Precision, recall, specificity, precision-recall F1 score, sensitivity-specificity F1 score, mean average precision, accuracy and kappa are applied as metrics.\\
    \vspace{0.15cm}
    \centering
    \begin{adjustbox}{width=0.85\textwidth}
        \begin{tabular}{l|ccccccccc|c}
        \hline
         \textbf{Model} & \textbf{Normal} & \textbf{dAMD} & \textbf{CSC} & \textbf{DR} & \textbf{GLC} & \textbf{MEM} & \textbf{MYO} & \textbf{RVO} & \textbf{wAMD} & \textbf{Average}\\ 
         \hline
        {Fundus} & 84.02 & 53.58 & 24.56 & 71.49 & 63.67 & \underline{43.33} & 63.54 & 80.77 & 34.74 & 57.74 \\
        {Fundus+Concept} & 84.74 & \underline{54.29} & 28.17 & \underline{71.95} & 64.67 & 36.07 & 64.75 & 78.80 & \underline{36.45} & 57.76 \\
        FFDM~\cite{wang2023fundus} & \underline{84.81} & 53.59 & \underline{36.48} & 71.72 & \underline{64.75} & 35.18 & \underline{66.18} & \underline{80.98} & \underline{36.45} & \underline{58.79} \\
        \rowcolor{cyan!60!blue!20}
        Ours & \textbf{84.90} & \textbf{54.71} & \textbf{43.64} & \textbf{72.08} & \textbf{66.91} & \textbf{48.39} & \textbf{67.90} & \textbf{82.83} & \textbf{36.87} & \textbf{62.02}\\
        \hline
        \end{tabular}
    \end{adjustbox}

    \vspace{0.1cm} 
    (b) Class-level results on the MultiEYE dataset with FLAIR backbone. Precision-recall F1 score is applied as the evaluation metric. \\
\end{table*}

\begin{table*}[t]
    \caption{Ablation of the two components, GPD and LCD, in our concept-assisted distillation on MultiEYE dataset. Class-level results are reported to show their respective influence on recognizing each disease. Precision-recall F1 score is applied as the evaluation metric.}
    \label{tab:ablation}
    \centering
    \begin{adjustbox}{width=0.8\textwidth}
        \begin{tabular}{cc|ccccccccc|c}
        \hline
        \multicolumn{2}{c|}{\textbf{Method}} & \multirow{2}{*}{\textbf{Normal}} & \multirow{2}{*}{\textbf{dAMD}} & \multirow{2}{*}{\textbf{CSC}} & \multirow{2}{*}{\textbf{DR}} & \multirow{2}{*}{\textbf{GLC}} & \multirow{2}{*}{\textbf{MEM}} & \multirow{2}{*}{\textbf{MYO}} & \multirow{2}{*}{\textbf{RVO}} & \multirow{2}{*}{\textbf{wAMD}} & \multirow{2}{*}{\textbf{Average}}\\ 
        \textbf{GPD} & \textbf{LCD} & & & & & & & & & \\
         \hline
        \ding{55} & \ding{55} & 84.74 & 54.29 & 28.17 & 71.95 & 64.67 & 36.07 & 64.75 & 78.80 & 36.45 & 57.76 \\
        \ding{51} & \ding{55} & 85.02 & 53.44 & 32.65 & 72.15 & 66.44 & 45.45 & 64.31 & \textbf{85.86} & 35.68 & 60.11 \\
        \ding{55} & \ding{51} & \textbf{85.65} & \textbf{55.59} & 40.00 & 70.99 & 65.29 & 43.33 & 62.88 & 84.31 & 36.45 & 60.50 \\
        \rowcolor{cyan!60!blue!20}
        \ding{51} & \ding{51} & 84.90 & 54.71 & \textbf{43.64} & \textbf{72.08} & \textbf{66.91} & \textbf{48.39} & \textbf{67.90} & 82.83 & \textbf{36.87} & \textbf{62.02}\\
        \hline
        \end{tabular}
    \end{adjustbox}
\end{table*}

\subsection{Performance on MultiEYE Dataset}
To evaluate our method, we make comparisons with several baselines. Unlike previous multi-modal learning approaches, no paired data are required during both training and test stages in our setting. Therefore, we construct the following baselines to compare with our proposed method.
\begin{itemize}
    \item \textbf{Fundus Model (Fundus)} includes a fundus feature extractor and a linear classifier.
    \item \textbf{Fundus Concept-Decoupled Model (Fundus+Concept)} is composed of a fixed text encoder, a trainable vision encoder and a concept classifier. The similarity between image features and concept embeddings is adopted as the input of the concept classifier.
    \item \textbf{FDDM}~\cite{wang2023fundus} proposes class prototype matching and class similarity alignment to distill disease knowledge. To align with our setting, we modify the knowledge flow direction to proceed from OCT modality to fundus modality.  
\end{itemize}

In our OCT-CoDA framework, we build upon VLMs to obtain the pre-trained weights of the image encoder and freeze the parameters of the text encoder. The image encoder is fine-tuned to align with the text embeddings. We apply different VLMs as backbones to evaluate our method, including CLIP~\cite{radford2021learning} and FLAIR~\cite{silva2023foundation}. The experimental results are reported in Table~\ref{tab:main-result} (a). 
As FLAIR is a well-versed VLM pre-trained on fundus dataset, preserving knowledge on ophthalmologic imaging and terminologies, its performance is relatively more competitive. We notify that fundus model and fundus concept-decoupled model show similar performance, but incorporating concepts can improve model interpretability (See in~\ref{sec:interpret}). 
Compared with the fundus model, OCT-CoDA improves the P-R F1 score by 2.77\% and 4.28\%, S-S F1 score by 1.63\% and 3.46\% on CLIP and FLAIR respectively. 
Besides, OCT-CoDA increases the performance in comparison with fundus concept-decoupled model by 2.22\% in P-R F1 score on CLIP backbone and 4.26\% on FLAIR backbone, indicating the effect of extra information from OCT B-scans.
Notably, OCT-CoDA outperforms the baseline FDDM in the FLAIR setting with 3.23\% performance gains in P-R F1 score, 5.21\% in S-S F1 score and 2.47\% in MAP, which demonstrates that the guidance of decoupled concepts could further enhance the efficacy of distillation.

To assess the efficacy of OCT-CoDA in diagnosing each disease, we present the class-level result on the MultiEYE dataset in Table~\ref{tab:main-result} (b). Comparing FFDM with single-modal methods, it can be noticed that distilling from OCT modality would enhance the fundus model in classifying most diseases, especially CSC and MYO. This could be attributed to the sensitivity of the OCT modality in detecting subtle retinal pathologies. However, FFDM struggles with reduced effectiveness in diagnosing certain diseases such as dAMD and MEM. A possible reason is that the undifferentiated distillation in FDDM might transfer noises, thus impeding the performance of the target modality.
In contrast, through decoupling the disease attributes and implementing distillation in a global-local manner, our approach successfully borrows extra information from the OCT modality and preserves the existing strengths of fundus images. Our proposed OCT-CoDA thereby surpasses the baselines across all categories.

\subsection{Analytical Studies}
\subsubsection{Effectiveness of GPD and LCD}
In Table ~\ref{tab:ablation}, we present an ablation study of our proposed components on the MultiEYE dataset to evaluate the effectiveness of Global Prototypical Distillation (GPD) and Local Contrastive Distillation (LCD). Our experimental result demonstrates that GPD alone improves the P-R F1 score by 2.35\% and LCD alone outperforms the baseline by 2.74\%. By combining these two methods, we achieve even higher performance gains of 4.26\%.
Also, the results infer the influence of GPD and LCD in detecting different diseases. GPD is superior in improving the classification performance in MEM and RVO, probably because the general features in OCT are more ideal for discovering these diseases than fundus photos. 
On the contrary, LCD benefits the detection of attributes that are already clearly visible in fundus photos, thus achieving higher scores in dAMD and wAMD. 

This demonstrates that our proposed GPD and LCD methods complementarily benefit the knowledge transfer process from the OCT modality to the fundus modality. While GPD encourages the generalized disease-specific knowledge to transfer from the OCT teacher model to the fundus student model, LCD helps the fundus model to maintain specific features instead of entirely depending on the OCT modality. By leveraging GPD and LCD in parallel, our method enables the fundus model to attend to the dominant features of the OCT modality while preserving its own advantageous characteristics, thus achieving a balanced result in each category.



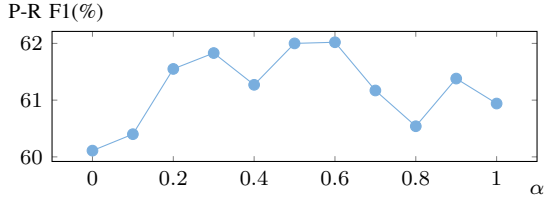
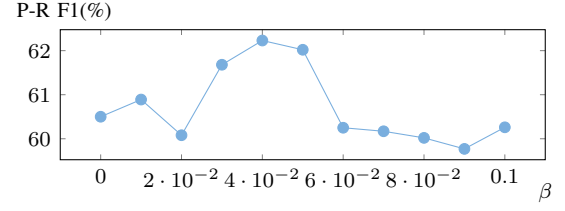
\begin{figure*}[t]
    \centering
    \begin{subfigure}{0.49\textwidth}
        \centering
        \begin{tikzpicture}
            \begin{axis}[
                xlabel=$\alpha$,
                ylabel=P-R F1(\%),
                tick align=inside,
                width=0.9\linewidth,  
                height=1.3in,
                tick style={major tick length=2pt},
                xlabel style={at={(axis description cs:1, 0.2)}, anchor=north},
                ylabel style={at={(axis description cs:0.2, 1)}, anchor=south, rotate=270},
                tick label style={font=\footnotesize},
                label style={font=\footnotesize}
                ]
                \addplot[mark=*, mark options={fill=cyan!60!blue!50}, cyan!60!blue!50] plot coordinates{
                (0, 60.11)
                (1e-1, 60.40)
                (2e-1, 61.55)
                (3e-1, 61.83)
                (4e-1, 61.27)
                (5e-1, 62.00)
                (6e-1, 62.02)
                (7e-1, 61.17)
                (8e-1, 60.54)
                (9e-1, 61.38)
                (1, 60.94)
                };
            \end{axis}
        \end{tikzpicture}
        \caption{$\alpha$ ranging from 0 to 1 and $\beta$ fixed to 0.05.}
    \end{subfigure}
    \begin{subfigure}{0.49\textwidth}
        \centering
        \begin{tikzpicture}
            \begin{axis}[
                xlabel=$\beta$,
                ylabel=P-R F1(\%),
                tick align=inside,
                width=0.9\linewidth,  
                height=1.3in,
                tick style={major tick length=2pt},
                xlabel style={at={(axis description cs:1, 0.2)}, anchor=north},
                ylabel style={at={(axis description cs:0.2, 1)}, anchor=south, rotate=270},
                tick label style={font=\footnotesize},
                label style={font=\footnotesize}
                ]
                \addplot[mark=*, mark options={fill=cyan!60!blue!50}, cyan!60!blue!50] plot coordinates{
                (0, 60.50)
                (1e-2, 60.89)
                (2e-2, 60.08)
                (3e-2, 61.68)
                (4e-2, 62.23)
                (5e-2, 62.02)
                (6e-2, 60.25)
                (7e-2, 60.17)
                (8e-2, 60.02)
                (9e-2, 59.77)
                (1e-1, 60.26)
                };
            \end{axis}
        \end{tikzpicture}
        \caption{$\beta$ ranging from 0 to 0.1 and $\alpha$ fixed to 0.6.}
    \end{subfigure}
    
    \caption{Analysis of the loss weight sensitivity on the MultiEYE dataset. We draw the curve of the Precision-Recall F1 score.}
    \label{fig:loss-weight}
\end{figure*}

\subsubsection{Sensitivity of Balancing Weights}
In Figure~\ref{fig:loss-weight}, we perform ablation experiments on the balancing loss weights $\alpha$ and $\beta$ for GPD and LCD. As shown in Figure~\ref{fig:loss-weight}, we observe that the final P-R F1 score follows a similar tendency with changes in both balancing weights. The distilled knowledge is insufficient when the value of $\alpha$ or $\beta$ is low. As the balancing weights increase, the P-R F1 score shows a general trend of increase and raises to 62.02\% when $\alpha$ and $\beta$ are set to $6e^{-1}$ and $5e^{-2}$ respectively. However, a higher value of loss weights can impede performance improvement, probably because redundant distilled knowledge would negatively impact the learning of the original data distribution with classification loss.


\subsubsection{Extension to Multi-Modal Setting on Paired Dataset}
\response{Considering that OCT scans are more effective for diagnosing retinal diseases and should be fully utilized if accessible, we extend our setting to scenarios where both modalities are provided for testing. Specifically, the enhanced fundus model and the pre-trained OCT model are applied for fundus and OCT inputs respectively. We average the prediction score of the pre-trained OCT model ($p_o$) and the fundus model ($p_f$), and then select the disease with the highest score as the final label. When paired data are accessible, we include the multi-modal fusion methods Concatenation~\cite{wang2019two}, Cross-Attention~\cite{praveen2021cross}, and Joint-Cross-Attention~\cite{praveen2022joint} as baselines.}

\begin{table}[t]
    \caption{\response{Performance on the MMC-AMD Paired Dataset. We report precision-recall F1 score, accuracy and kappa as metrics. "Train" and "Test" indicate the modality applied in each stage. "Extended" means that our setting is extended to the same scenario as multimodal models where both modalities are provided for test.}}
    \label{tab:paired-data}
    \centering
    \begin{adjustbox}{width=0.48\textwidth}
        \begin{tabular}{l|cc|cccc}
        \hline
         \textbf{Method} & \textbf{Train} & \textbf{Test} & \textbf{P-R F1} & \textbf{Accuracy} & \textbf{Kappa} \\ 
         \hline
        Fundus & Fundus & Fundus & 76.35 & 73.28 & 63.17 \\
        Fundus+Concept & Fundus & Fundus & 77.38 & 77.50 & 70.00 \\
        \hdashline
        FDDM~\cite{wang2023fundus} & Both & Fundus & 78.16 & 78.75 & 71.67 \\
        \rowcolor{cyan!60!blue!20}
        Ours & Both & Fundus & 81.62 & 82.50 & 76.67 \\ 
        \hline
        Two-Stream~\cite{wang2019two} & Both & Both & 84.95 & 83.21 & 77.00 \\ 
        Cross-Attention~\cite{praveen2021cross} & Both & Both & 88.73 & 94.30 & 84.95 \\ 
        Joint-Cross-Attention~\cite{praveen2022joint} & Both & Both & 85.92 & 90.37 & 80.81 \\ 
        \rowcolor{cyan!60!blue!20}
        Ours (Extended) & Both & Both & \textbf{94.86} & \textbf{94.94} & \textbf{93.24} \\
        \hline
        \end{tabular}
    \end{adjustbox}
\end{table}

\begin{table}[t]
    \caption{\response{Comparison of model performance between seen and unseen devices. $\star$ means that the device is not included in the training set.}}
    \label{tab:unseen-device}
    \centering
    \begin{adjustbox}{width=0.48\textwidth}
        \begin{tabular}{l|cccccc}
        \hline
         \textbf{Dataset} & \textbf{Precision} & \textbf{Recall} & \textbf{P-R F1} & \textbf{S-S F1} & \textbf{Accuracy} & \textbf{MAP} \\ 
         \hline
         \textbf{Messidor-2}$\star$ & 79.19 & 76.00 & 76.67 & 73.44 & 78.33 & 82.43 \\
          \textbf{OIA-DDR}$\star$ & 82.09 & 79.90 & 79.95 & 78.12 & 80.42 & 87.47 \\ 
         \textbf{EYEPACS} & 79.29 & 78.98 & 79.13 & 78.98 & 77.66 & 63.97 \\
         \hline
        \end{tabular}
    \end{adjustbox}
\end{table}

\begin{table}[t]
    \caption{Results with different concept generation approaches on the MultiEYE dataset. In the generation process, the total number of concepts for each class is set to 10. The FLAIR model is used as the backbone.}
    \label{tab:concept-1}
    \centering
    \begin{adjustbox}{width=0.48\textwidth}
        \begin{tabular}{l|cccccc}
        \hline
         \textbf{Generation} & \textbf{Precision} & \textbf{Recall} & \textbf{P-R F1}& \textbf{S-S F1} & \textbf{MAP} & \textbf{Kappa} \\ 
         \hline
         \textsc{Expert} & 67.12 & 61.93 & 61.42 & 71.30 & 63.52 & \textbf{60.27} \\
         \textsc{GPT-3.5} & 67.01 & 61.93 & 61.45 & 71.67 & 63.27 & 58.71 \\ 
         \textsc{GPT-4} & 66.60 & \textbf{63.45} & 61.49 & 71.85 & 64.23 & 58.99 \\
         \textsc{GPT-4+CoT} & \textbf{67.15} & 62.61 & \textbf{62.02} & \textbf{72.25} & \textbf{64.37} & 59.67 \\
         \hline
        \end{tabular}
    \end{adjustbox}
\end{table}

\begin{figure}[t]
	\centering
	\includegraphics[width=0.49\textwidth]{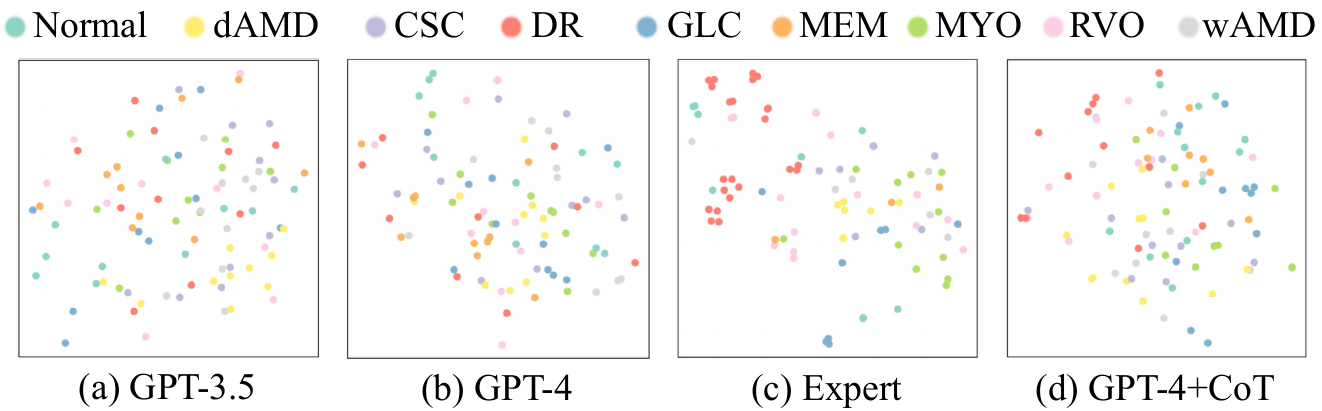}
	\caption{Embedding distribution of concepts from different generation methods.}
	\label{fig:concept-distribution}
\end{figure}

To compare our method with the multi-modal approaches, we apply a paired dataset, MMC-AMD~\cite{wang2022learning}, which contains \response{1,093 fundus images and 1,288 OCT B-scans} of four classes including normal, dry AMD, polypoidal choroidal vasculopathy (PCV), and wet AMD. In this dataset, we resize all the images into 224$\times$224 to be consistent with the setting in Two-Stream CNN~\cite{wang2019two}. The experimental results are shown in Table~\ref{tab:paired-data}. It is obvious that given only fundus images at the test stage, our method still outperforms single fundus modal baselines and FDDM comparatively by 5.27\% and 3.46\% in F1 score, benefitting from the distilled OCT knowledge. 
\response{This demonstrates the robustness} \response{of our method even when the stronger OCT modality is not available for testing.
By combining our enhanced fundus model with pre-trained OCT model to tackle the multi-modal inputs, we can achieve better performance than the multi-modal fusion methods. In conclusion, our extended method is flexible to integrate explicit information from fundus and OCT modalities when both are provided at the test stage. If only a single fundus modality is applicable, our proposed OCT-CoDA is still capable of using the implicit OCT knowledge to improve the performance.}

\begin{figure*}[t]
	\centering
	\includegraphics[width=0.95\textwidth]{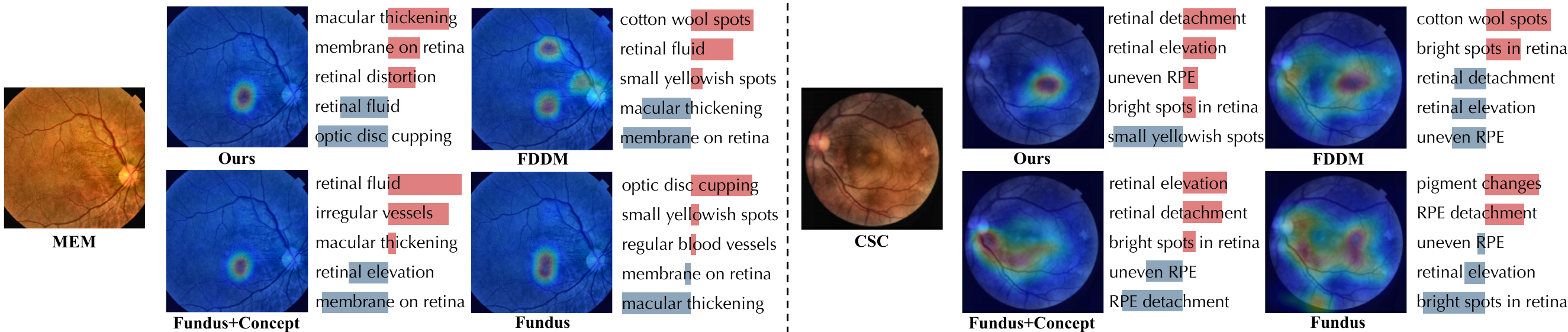}
	\caption{Concept visualization on MultiEYE dataset. For each case, we show the heat map to reflect the activated areas. Besides, we select five concepts and display the normalized similarity score with red or blue bars, indicating the score is above or below the average score of all the concepts for this sample. Here, we truncate the detailed description of color, shape and location in each concept due to limited space.}
	\label{fig:vis}
\end{figure*}

\begin{table}[t]
    \caption{Results with different concept selection approaches on the MultiEYE dataset. We select 10 concepts for each class from the large concept pool generated by GPT-4 with the CoT prompt. The FLAIR model is applied as the backbone.}
    \label{tab:concept-2}
    \centering
    \begin{adjustbox}{width=0.5\textwidth}
        \begin{tabular}{l|cccccc}
        \hline
         \textbf{Selection} & \textbf{Precision} & \textbf{Recall} & \textbf{F1\_PR} & \textbf{F1\_SS} & \textbf{MAP} & \textbf{Kappa} \\ 
         \hline
         \textsc{None} & 67.15 & 62.61 & \textbf{62.02} & 72.25 & \textbf{64.37} & 59.67 \\
         \textsc{Random} & 64.49 & 61.53 & 59.99 & 71.08 & 63.71 & 59.53 \\
         \textsc{SVD} & 62.98 & 62.28 & 59.58 & 71.69 & 61.53 & 59.11 \\
         \textsc{K-means} & 66.27 & 62.43 & 61.18 & 71.97 & 63.30 & 59.40 \\
         \textsc{Similarity} & \textbf{68.32} & 61.29 & 61.77 & 70.92 & 63.58 & 58.80 \\
         \textsc{Submodular} & 63.09 & \textbf{64.02} & 61.02 & \textbf{73.58} & 63.81 & \textbf{59.71} \\ 
         \hline
        \end{tabular}
    \end{adjustbox}
\end{table}

\begin{table}[t]
    \caption{Results with different concept pool scales on the MultiEYE dataset. All the concepts are directly generated by GPT-4 with CoT prompts. The FLAIR model is used as the backbone.}
    \label{tab:concept-3}
    \centering
    \begin{adjustbox}{width=0.48\textwidth}
        \begin{tabular}{c|cccccc}
        \hline
         \textbf{\makecell{Concepts\\Per Class}} & \textbf{Precision} & \textbf{Recall} & \textbf{P-R F1} & \textbf{S-S F1} & \textbf{MAP} & \textbf{Kappa} \\ 
         \hline
         5 & 63.52 & 61.04 & 59.09 & 70.12 & 58.82 & 58.07 \\
         10 & \textbf{67.15} & \textbf{62.61} & \textbf{62.02} & \textbf{72.25} & \textbf{64.37} & 59.67 \\
         20 & 64.66 & 62.43 & 61.08 & 71.64 & 62.28 & \textbf{60.28} \\
         50 & 64.88 & 60.11 & 59.64 & 69.65 & 64.21 & 60.00 \\
         \hline
        \end{tabular}
    \end{adjustbox}
\end{table}

\begin{figure}[t]
	\centering
	\includegraphics[width=0.5\textwidth]{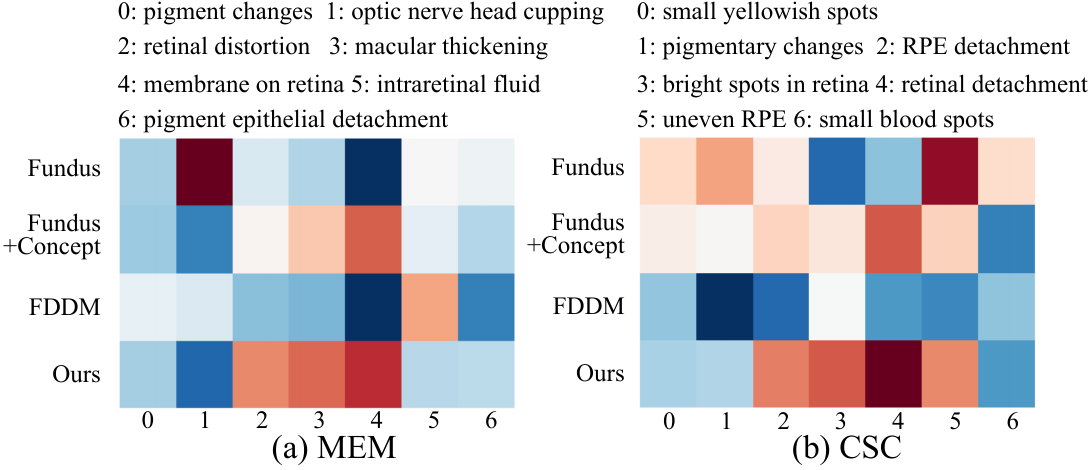}
	\caption{The average concept similarity score for all the images of MEM and CSC. We select 7 concepts for demonstration. (a) For MEM, concept 2, 3, 4 can be present. (b) For CSC, concept 2, 3, 4, 5 are likely to appear.}
	\label{fig:vis2}
\end{figure}

\subsubsection{Impact of domain shift on different imaging devices.} \response{When constructing the fundus dataset, we included Messidor-2 and OIA-DDR exclusively for validation and test, which can indicate the ability of our model to adapt to a new device, Topcon TRC NW6 fundus camera. Both Messidor-2 and OIA-DDR datasets include normal and diabetic retinopathy (DR) labels, sharing the same label space as the EYEPACS dataset. To investigate the effect of different imaging devices on model performance, we separately evaluated our model on these three datasets. It can be inferred from Re-Table II that the result on these two datasets is comparable to that of EYEPACS, despite the images coming from an unseen device. This demonstrates the robustness of our model in handling images collected from different devices, i.e., under conditions of domain shifts.}

\subsubsection{Analysis of Concept Generation Approaches}
We compare different concept generation approaches: (1) \textsc{Expert}: we retrieve disease-related concepts from clinical literature and public resources including EyeWiki\footnote{https://eyewiki.org/Main\_Page} and MSD Manual\footnote{https://www.msdmanuals.com/professional}. (2) \textsc{GPT-3.5}/\textsc{GPT-4}: we directly use GPT-3.5~\cite{liu2023chatgpt} and GPT-4~\cite{openai2023gpt4} to obtain the required concepts, with prompts like ``\textit{Describe the typical symptoms of \{disease name\} in fundus or OCT images}". 
(3) \textsc{GPT-4+CoT}: as described in Section~\ref{sec:gpt}, we acquire concepts from GPT-4 with CoT prompts. 

From Table~\ref{tab:concept-1}, it is evident that employing LLMs for concept generation does not significantly hinder performance compared with utilizing attributes derived from expert knowledge. This indicates that GPT Series LLMs possess ample information on eye diseases, which can be leveraged to improve model training and facilitate cross-modal knowledge transfer. 
Additionally, incorporating the CoT approach into prompt design could further yield more informative concepts, as suggested by the improvements in P-R F1, S-S F1, and MAP metrics.
To further analyze different concept generation methods, we display the distribution of concept embeddings in Figure~\ref{fig:concept-distribution}. Since different diseases might exhibit similar appearances, concepts generated with different disease prompts can be close in the embedding space.
We can observe that the concepts directly generated by GPT-3.5 and GPT-4 tend to scatter spatially, while those from the same class are more concentrated for expert and GPT-4+CoT. It infers that refined prompts stimulate GPT to think like humans and generate rich semantics.


\subsubsection{Analysis of Concept Selection Approaches}
As shown in Table~\ref{tab:concept-2}, we explore different concept selection methods to filter $K$ concepts: (1) \textsc{None}: we directly use LLM to generate the concepts; (2) \textsc{Random}: we randomly sample a subset of $K$ concepts from the concept pool; (3) \textsc{SVD}: we employ Singular Value Decomposition (SVD) on concept embeddings to identify the top $K$ vectors and find concepts with the highest similarity to these significant vectors. (4) \textsc{K-means}: we perform K-means clustering on concept embeddings, and obtain $K$ attributes that are closest to each cluster center. (5) \textsc{Similarity}: we select the top $K$ concepts ranked by the averaged similarity score across all images. (6) \textsc{Submodular}: we follow~\cite{yang2023language} to utilize submodular optimization to greedily select a subset of $K$ concepts, thereby maximizing both discriminability and diversity. In the selection process, all the features are calculated with the pre-trained weights of the FLAIR model.

From the experimental results, we notice that random and SVD-based selection are less competitive compared with other methods, which manifests their weakness in recognizing informative concepts. Conversely, other selection algorithms generally yield substantial results, achieving over 61\% in P-R F1 score. Moreover, directly applying GPT-4 to generate an equivalent number of disease attributes does not impact classification outcomes even without selection. This infers that GPT-4 is prone to generate principal concepts provided an appropriate total quantity.

\subsubsection{Analysis of Concept Pool Scale}
In Table~\ref{tab:concept-3}, we compare the performance of different concept pool scales with concepts ranging from 5 to 50 per class, directly generated by LLM. As the concept pool scale enlarges, the classification performance improves. The best P-R F1, S-S F1 and MAP are reached when 10 concepts are generated for each disease. When the concept pool scale continues increasing, the classification performance drops apparently. We attribute the phenomenon to the reason that a small concept pool could not sufficiently describe disease characters, while a too large concept pool may include repetitive and redundant information.

\subsubsection{Concept Intepretability and Visualization}
\label{sec:interpret}
Our proposed conceptual distillation approach can also interpret what knowledge is transferred from the OCT teacher model to enhance the fundus student model. In Figure~\ref{fig:vis}, we take two diseases, MEM and CSC, as examples and demonstrate case studies. 
It can be observed that our method not only relates the disease attributes of abnormal fundus photos with corresponding lesion regions, but also transfers the OCT knowledge by increasing the similarity with OCT dominant characters. For example, compared with FDDM, our method manages to extract correct concepts correlated with the activated region, which proves that useful knowledge is filtered and distilled from the OCT modality. As for the fundus concept-decoupled model, it either fails to recognize concepts more obvious in OCT modality like ``retinal elevation" in the MEM sample, or cannot associate with the correct macula area as shown in the CSC sample. Conversely, after leveraging our OCT-CoDA method, we increase the similarity score of disease attributes that are clearer in the OCT modality, and align them with lesions.

Moreover, we compute the averaged concept similarity score for all the images of MEM and CSC, as demonstrated in Figure~\ref{fig:vis2}. It can be noticed that our method not only gains a closer relation between diseases and retinal layer attributes, but also preserves fundus dominant features like ``bright spots in retina".
This indicates that our method proposes an interpretable distillation by enhancing OCT dominant features in the fundus modality, and improves the diagnostic performance based on fundus photos.

\section{Discussion}
\label{sec:discussion}

Automatic retinal disease classification models have attempted to jointly adopt fundus photos and OCT images for diagnosis. Although satisfactory accomplishments have been witnessed, these approaches are mainly limited by the requirement for paired modalities and the inexplicable knowledge flow between modalities. In this work, we construct a novel benchmark for OCT-enhanced fundus image screening and propose an OCT-assisted Conceptual Distillation Approach (OCT-CoDA) to augment the fundus modality with the more reliable OCT modality. We guide the knowledge transfer process with fine-grained disease attributes generated from LLM, thus making knowledge sharing more controllable and explainable. In the inference stage, only the cheaper and more affordable fundus images are required for the final prediction. 

Despite the outstanding performance, our method has certain limitations. 
\response{Firstly, our model is developed under the assumption that the diseases being tested are seen diseases, leaving unseen diseases unresolved. To effectively manage unseen diseases and to enhance our model's adaptability to a broader range of input images, we plan to design a linear binary classification layer beyond the existing disease classification layer, allowing the network to classify the likelihood of diseases being seen or unseen.}
Additionally, our proposed OCT-CoDA relies on the ophthalmology knowledge preserved in the pre-trained foundation model. The alignment between image features and disease-related concepts impacts the performance of our approach, see in Table~\ref{tab:main-result}.
Furthermore, we expect future research to assemble more modalities, such as Fundus Fluorescein Angiography (FFA) and Optical Coherence Tomography Angiography (OCTA). Consequently, superior knowledge from various modalities could be gathered to further improve the diagnostic performance. 

\section{Conclusion}
\label{sec:conclusion}
\response{In our project, we introduce a novel approach to ``OCT-enhanced retinal disease classification from fundus photos," utilizing unpaired multi-modal OCT and fundus data (data that are not from the same patients but share the same label space), with only fundus photographs available for testing. We first construct a large multi-modal, multi-class benchmark called MultiEYE and propose an OCT-assisted Conceptual Distillation Approach (OCT-CoDA), which aims to distill disease-related knowledge from OCT into the fundus model during training. Our core idea is to leverage the relation of distinct appearances between two modalities into the fundus model during training, thereby making it capable to imply the typical features concerning the depth and thickness of the retina despite the OCT modality not being accessible.   Our experimental analysis shows that our methodology significantly outperforms existing approaches. Ablation studies and visualizations further illustrate the interpretability of our model in the knowledge transfer process between modalities. We believe this research is valuable as it is the first to demonstrate the utility of unpaired OCT and fundus data in enhancing model performance when only fundus images are available for testing. It offers an effective approach to strengthen fundus models using available OCT datasets.}

\bibliographystyle{IEEEtran}
\bibliography{refs}

\end{document}